\documentclass[10pt,twocolumn,letterpaper]{article}

\usepackage{cvpr}
\usepackage{times}
\usepackage{epsfig}
\usepackage{graphicx}
\usepackage{amsmath}
\usepackage{amssymb}
\usepackage{enumitem} 
\usepackage{booktabs}
\usepackage{multirow}
\usepackage{bigstrut}
\usepackage{comment}
\usepackage[dvipsnames]{xcolor}

\usepackage[pagebackref=true,breaklinks=true,letterpaper=true,colorlinks,bookmarks=false]{hyperref}

\cvprfinalcopy 

\def\cvprPaperID{677} 

\ifcvprfinal\fi
\begin{document}
	
	\title{Densely Semantically Aligned Person Re-Identification}
	\author{Zhizheng Zhang{$^1$}\thanks{This work is done when Zhizheng Zhang is an intern at MSRA.} \qquad 
	Cuiling Lan$^2$\thanks{Corresponding author}
	\qquad 
	Wenjun Zeng$^2$ 
	\qquad 
	Zhibo Chen$^{1\dagger}$
	\and 
	\normalsize
		$^1$University of Science and Technology of China \qquad $^2$Microsoft Research Asia\\
		{\tt\small zhizheng@mail.ustc.edu.cn} \qquad {\tt\small \{culan, wezeng\}@microsoft.com} \qquad {\tt\small chenzhibo@ustc.edu.cn}
	}
	
	\maketitle
	\thispagestyle{empty}
	
	\begin{abstract}
		
		We propose a densely semantically aligned person re-identification framework. It fundamentally addresses the body misalignment problem caused by pose/viewpoint variations, imperfect person detection, occlusion, {\it{etc}}. By leveraging the estimation of the dense semantics of a person image, we construct a set of densely semantically aligned part images (DSAP-images), where the same spatial positions have the same semantics across different images. We design a two-stream network that consists of a main full image stream (MF-Stream) and a densely semantically-aligned guiding stream (DSAG-Stream). The DSAG-Stream, with the DSAP-images as input, acts as a regulator to guide the MF-Stream to learn densely semantically aligned features from the original image. In the inference, the DSAG-Stream is discarded and only the MF-Stream is needed, which makes the inference system computationally efficient and robust. To the best of our knowledge, we are the first to make use of fine grained semantics to address the misalignment problems for re-ID. Our method achieves rank-1 accuracy of 78.9\% (new protocol) on the CUHK03 dataset, 90.4\% on the CUHK01 dataset, and 95.7\% on the Market1501 dataset, outperforming state-of-the-art methods. 
	\end{abstract}
	
	\section{Introduction}
	
	Person re-identification (re-ID) aims to match a specific person across multiple camera views or in different occasions from the same camera view. It facilitates many important applications, such as cross-camera tracking \cite{wang2013intelligent}.
	
\begin{figure}[t]
		\begin{center}
			\includegraphics[width=1\linewidth]{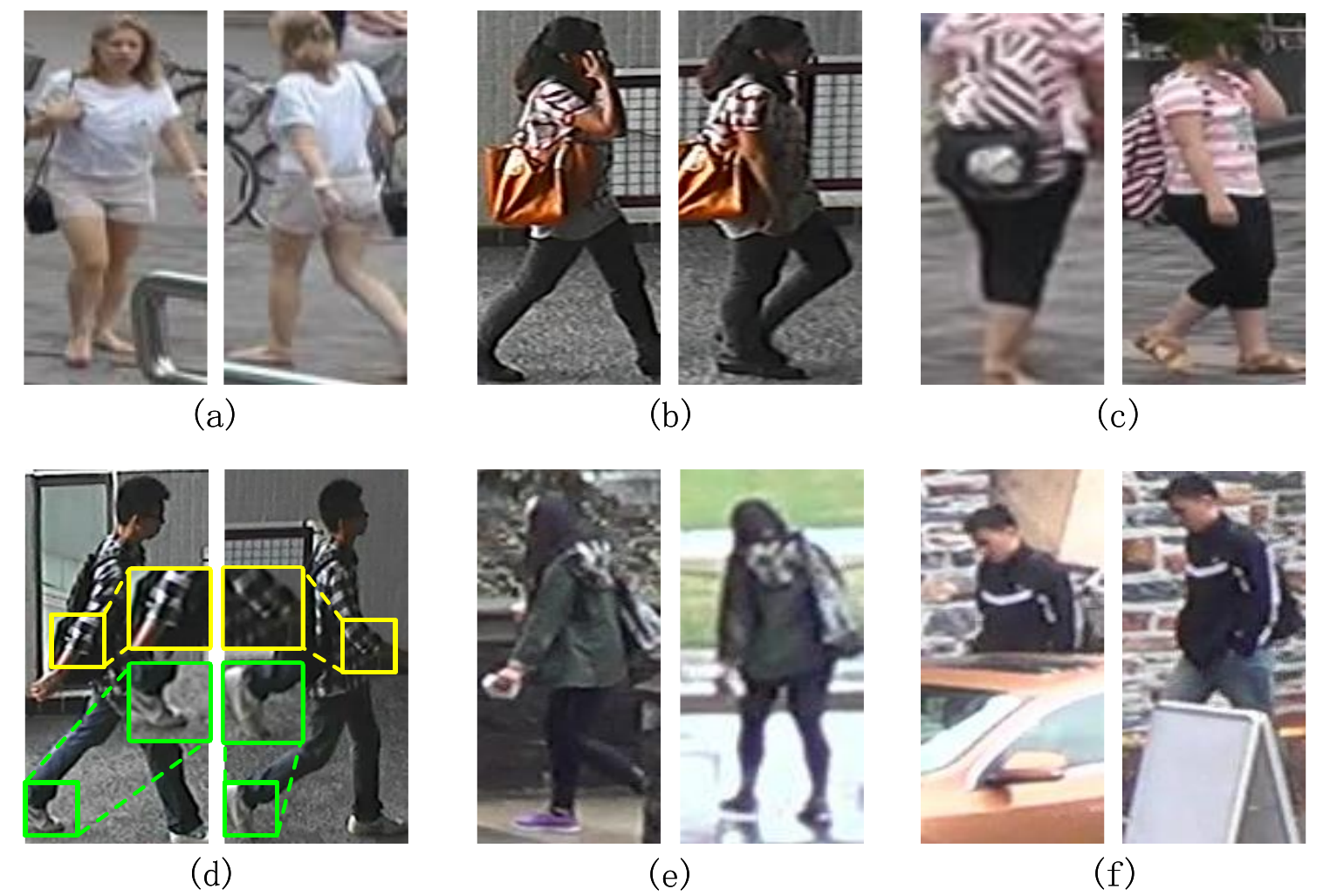}
		\end{center}
		\caption{Examples to illustrate the challenges of spatial misalignment in person re-ID caused by (a) different camera viewpoints, (b) different poses, (c) imperfect person detection, (d) misalignment within a local part, (e) cluttered background, (f) occlusion.}
		\label{fig:examples}
	\end{figure}

	This task is challenging due to large variations on person pose and viewpoint, imperfect person detection, cluttered background, occlusion, and lighting differences, {\it{etc}}. Many of these factors result in spatial misalignment of the human body as shown in Fig.~\ref{fig:examples}, where the same spatial positions do not correspond to the same semantics. The misalignment is one of the key challenges~\cite{shen2015person,su2017pose,zheng2017pose,varior2016siamese,zhang2017alignedreid,suh2018part,zheng2018pedestrian}, which compromises performance.
	
	Some paradigms employ the convolutional neural networks to learn global feature representation in an end-to-end manner \cite{ahmed2015improved,wu2016personnet,geng2016deep,chen2017multi,almazan2018re}. However, the capability of the global representations is limited by: 1) the lack of emphasis on local differences \cite{zhang2017alignedreid}, and 2) the absence of any explicit mechanism to tackle the misalignment \cite{almazan2018re}. 
	
	

	In recent years, many efforts have been made to alleviate these problems \cite{su2017pose,zheng2017pose,varior2016siamese,zhang2017alignedreid,suh2018part,li2018harmonious}. To make the features focus on some local details, some works make a straightforward partition of the person image into a few fixed rigid parts ({\it{e.g.}}, horizontal stripes) and learn detailed local features \cite{cheng2016person,varior2016siamese,li2017person,bai2017deep,sun2017beyond,wang2018learning}. However, such a partition cannot well align the human body parts. Some works have attempted the use of pose (which identifies different types of parts, {\it{e.g.}}, head, arm, {\it{etc.}}) to localize body parts for learning part-aligned features \cite{yao2017deep,li2017learning,zhao2017spindle,su2017pose,zheng2017pose,wei2017glad}. However, the body part alignment based on pose is too coarse to have satisfactory alignment. As shown in Fig.~\ref{fig:examples}~(d), even for the same type of parts, there is still spatial misalignment within the parts, where the human semantics are different for the same spatial positions. It becomes critical to design an architecture which enables the efficient learning of densely semantically aligned features for re-ID.
	
	\begin{figure}[t]
		\begin{center}
			\includegraphics[width=1\linewidth]{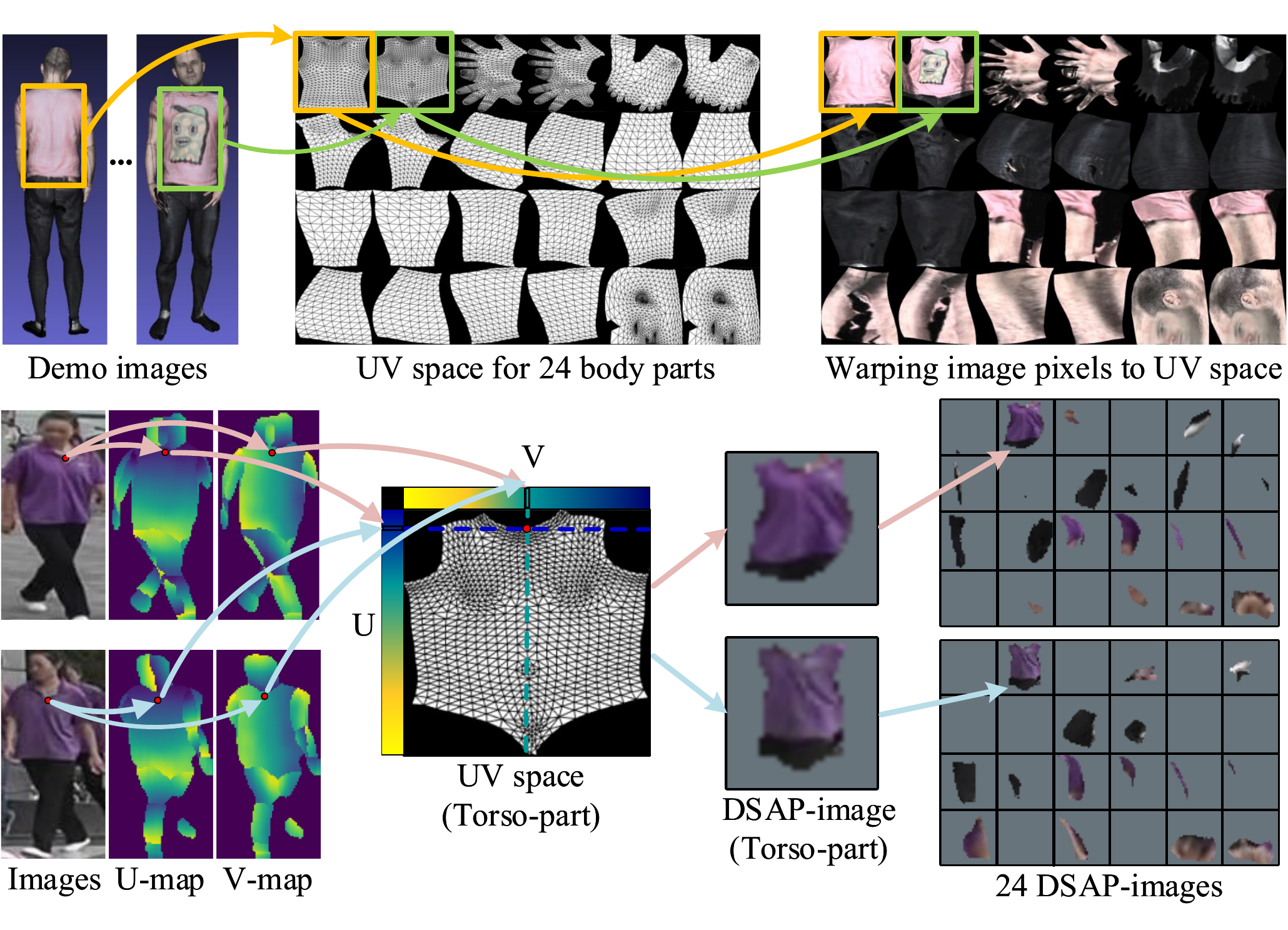}
		\end{center}
		\vspace{-3mm}
		\caption{Illustration of the dense correspondences between a 2D person image and a surface-based canonical representation in the UV space. The person surface is partitioned into 24 body regions. Each region can be warped to a DSAP-image and the fine-grained (dense) semantics are spatially aligned for different person images.}
		\vspace{-3mm}
		\label{fig:2d-3d}
	\end{figure}
	
	In this paper, we propose a novel densely semantically aligned person re-ID framework, which fundamentally enables fine-grained semantic alignment and semantically aligned feature learning in re-ID. 
	
	First, we propose performing dense semantic alignment of the human body on a canonical space to address the misalignment challenges in person re-ID. We are inspired by the dense semantics estimation work of DensePose~\cite{guler2018densepose}, which is capable of predicting the fine-grained pixel-level semantics of a person. Different from pose with only a limited number of coarse key joints, dense semantics establishes dense correspondences between a 2D person image and a 3D surface-based canonical representation of the human body \cite{guler2018densepose,guler2017densereg}. As illustrated in Fig.~\ref{fig:2d-3d}, the 3D surface of a person is segmented into 24 semantic body regions. Within a region, the semantics of each position is identified by a two-dimensional UV coordinate. Based on the estimated dense semantics in terms of UV coordinate values (on the U,V map), the original input image is warped to 24 \emph{densely semantically aligned part images} (DSAP-images) in the UV space. In this way, person images with different viewpoints, poses, and backgrounds are  semantically well aligned. Thus, such representation has the inherent merits for addressing misalignment challenges. Note that not only the coarse body part regions are aligned, but also the contents within a part are densely aligned at the pixel level.

	Second, we propose a new framework intending to fully exploit the densely semantically aligned representations for person re-ID. For dense semantics estimation, since the person in a 2D image is a projection from a 3D person, nearly half of the 3D surface is invisible and thus cannot be detected from the 2D image (see the examples of the 24 DSAP-images in  Fig.~\ref{fig:2d-3d}, where many of the DSAP-images do not have valid information). Besides, there are usually estimation errors, including missing detection, especially on the images of the re-ID dataset which usually have low resolution and blurring artifacts. It remains challenging to design an effective network to fully exploit the semantically aligned information as there are loss of information and noise there.

	In our design, we leverage the densely semantically aligned information to drive the main network to learn semantically aligned features from the original image. As shown in Fig.~\ref{fig:flowchat-learning}, our network consists of a main full image stream (\emph{MF-Stream}) and a densely semantically aligned guiding stream (\emph{DSAG-Stream}). For the \emph{MF-Stream}, the full image is taken as the input. For the \emph{DSAG-Stream}, the 24 DSAP-images obtained from the dense semantic alignment module are taken as the input. Rather than making the features of the two streams both have re-ID ability, the \emph{DSAG-Stream} acts as a regulator to guide the \emph{MF-Stream} to learn semantically aligned features. We achieve this by element-wise fusing of the \emph{MF-Stream} features and the \emph{DSAG-Stream} features, with supervisions added on the fused features. End-to-end joint training enables the interaction and joint optimization of the two streams.

	In summary, we have made three main contributions.
	
	\begin{itemize}[noitemsep,nolistsep,leftmargin=*] 
		
		\item We propose making use of dense semantic alignment for person re-ID, addressing the misalignment challenges. 
		
		\item A densely semantically aligned deep learning based framework is proposed for person re-ID. To the best of our knowledge, our proposed framework is the first one to make use of fine grained semantics to address the misalignment problems for effective person re-ID.
		We propose an effective fusion and supervision design to facilitate semantically aligned feature learning. It enables the interaction between the \emph{DSAG-Stream} and the \emph{MF-Stream} during the learning process. This greatly enhances the power of the \emph{MF-Stream} even though its input images are not semantically aligned.
		
		\item The \emph{DSAG-Stream}, as a regulator, can be removed during the inference without sacrificing the performance. This also removes the dependency on the performance of the dense semantics estimator during interference, making the inference model more computationally efficient and robust to dense semantics estimation errors.  
	\end{itemize}
	
	We perform extensive ablation studies and the experimental results demonstrate that our proposed architecture with the dense semantic alignment are very powerful. We achieve state-of-the art performance on the Market-1501, CUHK03, and CUHK01 datasets, and competitive performance on DukeMTMC-reID. On the CUHK03 dataset, our performance significantly outperforms the previous methods, by at least \textbf{+10.9\%/+7.8\%} in Rank-1/mAP accuracy. 

	\section{Related Work}
	
	\textbf{Body part/pose-aligned approaches.} Spatial misalignment is ubiquitous and is one of the key challenges in re-ID. In the early works, some patch-based methods perform patch-level matching to address patch-wise misalignment \cite{oreifej2010human,zhao2013unsupervised,zhao2014learning}. To avoid mismatched patches with similar appearances \cite{shen2015person}, human semantics of part/pose is introduced so that the similarity matching is performed between the semantically corresponding parts \cite{cheng2011custom,xu2013human}.
	In recent years, human semantics in terms of part/pose is widely used to localize body parts for part-aligned deep feature learning and matching~\cite{yao2017deep,li2017learning,zhao2017spindle,su2017pose,zheng2017pose,wei2017glad}. In \cite{wei2017glad}, body poses/parts are first detected and deep neural networks are designed for representation learning on both the local parts and global region. Some works rely on constrained attention selection mechanisms from human mask/part/pose to implicitly calibrate misaligned images~\cite{song2018mask,qi2018maskreid,xu2018attention,kalayeh2018human,suh2018part}. 

	All the above works aim to address misalignment at the coarse body part level. However, there is still misalignment within each part. Our work intends to fundamentally address the misalignment problem. It differs from previous works in three main aspects. First, our approach intends to fully exploit the fine-grained semantically aligned representations. Second, we leverage the semantically aligned representations, which play the role of regulators, to guide the semantic feature learning from the original image. Third, during inference, we do not need the \emph{DSAG-Stream}, making our model computationally efficient and robust.
	
	
	\textbf{Local and global based approaches.} Many approaches make use of both the global and local feature to simultaneously exploit their advantages \cite{bai2017deep,wang2018learning,li2017person,wei2017glad,su2017pose,zhao2017spindle,zhang2017alignedreid}. Global features learned from the full image intend to capture the most discriminative clues of appearance but may fail to capture discriminative local details. Thus, part-based features are exploited as a remedy. Wang {\it{et al.}} design a multiple granularity network, which consists of one branch for global features and two branches for local feature representations \cite{wang2018learning}. In \cite{zhang2017alignedreid}, the image feature map is rigidly divided into local stripes and a shortest path loss is introduced to align local stripes. This aids the global feature learning by means of sharing weights of the backbone network. However, the alignment is still too coarse without considering person dense semantics.
    We leverage the densely semantically aligned representation to \emph{guide} the learning of  both the global features and part-aware features.

	\textbf{Approaches based on joint multi-loss learning.} Zheng {\it{et al.}} suggest that person re-ID lies in between image classification and instance retrieval \cite{zheng2016person}. The classification task and ranking task are complementary to each other. Recently, some approaches \cite{wu2016personnet,liu2017end,chen2017multi,wang2018learning} optimize the network simultaneously with both classification loss and ranking loss, {\it{e.g.}}, triplet loss \cite{schroff2015facenet,hermans2017defense}. Similarly, we leverage the complementary advantages of the two tasks.
	
	
	
	
	\section{Densely Semantically Aligned Person Re-ID}
	
\begin{figure*}[th]
		\begin{center}
			\vspace{-5mm}
			\includegraphics[width=1\linewidth]{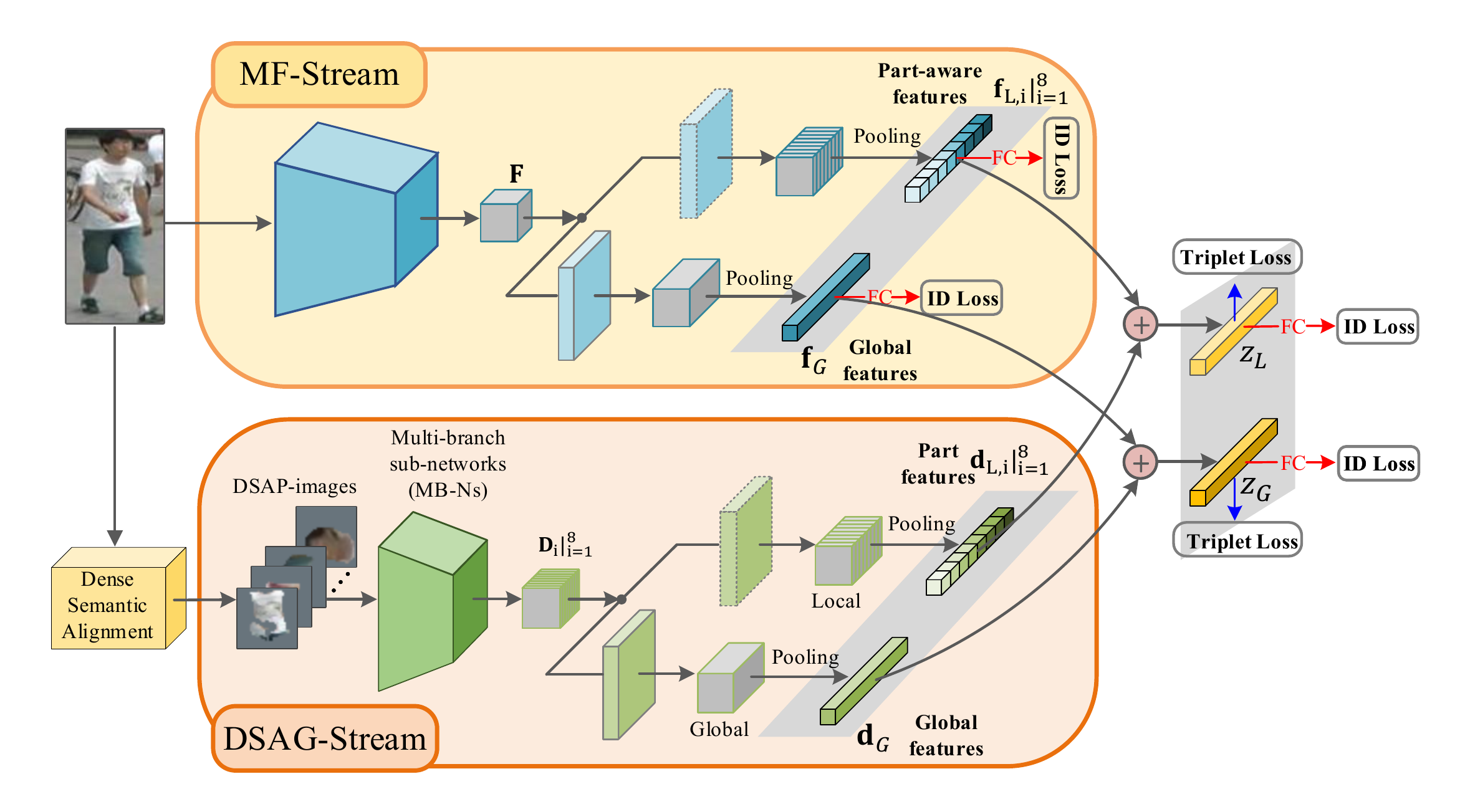}
		\end{center}
		\vspace{-6mm}
		\caption{Flowchat of the proposed densely semantically aligned person re-ID (DSA-reID). It consists of two streams: \emph{MF-Stream} and \emph{DSAG-Stream}, which are jointly trained through our fusion and supervision design. The \emph{DSAG-Stream}, with the input DSAP-images that are densely semantically aligned, plays the role of a regulator which facilitates the joint optimization of the entire network. In the inference, to be computationally efficient, the \emph{DSAG-Stream} is discarded. The global features $\mathbf{f}_G$ and part-aware features $\mathbf{f}_L = \mathbf{f}_{L,i}|_{i=1}^8$ are used as the final features for re-ID. They are simultaneously exploited to make use of the global information and local detailed information.}
		\label{fig:flowchat-learning}
	\end{figure*}

	We propose a new framework aiming to fully exploit the densely semantically aligned representations for robust person re-ID. Fig.~\ref{fig:flowchat-learning} shows the flowchat. The network consists of two streams: the main full image stream (\emph{MF-Stream}), and the densely semantically aligned guiding stream (\emph{DSAG-Stream}). Based on the dense semantic alignment module, from the input person image, we construct 24 densely semantically aligned part images (DSAP-images) as the input to the \emph{DSAG-Stream}. Having the merits of being semantically aligned, the \emph{DSAG-Stream} acts as a regulator to regularize the feature learning of the \emph{MF-Stream} from the original image, through our fusion and loss designs. The entire network is trained in an end-to-end manner. We discuss the details in the following subsections.

	\subsection{Construction of DSAP-images}
	\label{subsec:SAPIC}
	
	Dense semantics annotation/estimation on 2D images \cite{guler2018densepose} establishes dense correspondences from 2D images to the human body surface. Each position on the surface has a different semantic meaning, which can be parameterized/represented by a two-dimensional UV coordinate value \cite{guler2017densereg,guler2018densepose}. The same UV coordinate value corresponds to the same semantics. Thus, in the UV space, the dense semantics are inherently aligned. 
	
	For the dense semantic alignment module, the original RGB image is warped to the representation in UV space to obtain 24 DSAP-images based on the estimated dense semantics. 
	
	

	
	
	\textbf{Dense semantics estimation.} We adopt the off-the-shelf DensePose model (trained on the DensePose-COCO dataset) \cite{guler2018densepose} to estimate the dense semantics of a 2D image. It segments a person to 24 surface-based body part regions. For each detected body part, the semantics for each pixel is provided in terms of a coordinate value (u,v) in the UV space, where u,v $\in$[0,1]. Please refer to \cite{guler2018densepose} for more details.
	
	
	\textbf{Warping.} For the $i$-th body part region, as illustrated in Fig.~\ref{fig:2d-3d}, based on the semantics, the pixel values on the person can be warped onto a DSAP-image of size S$\times$S in the deformation-free UV space, where the rows and columns represent the U and V, respectively. The DSAP-images are initialized by the mean values of images before the warping. Note that the background and not detected semantic positions are not warped. We simply copy the pixel value (r,g,b) of the body parts with its semantics estimated as (u,v) to the ($\lfloor u \times S \rfloor$, $\lfloor v \times S \rfloor$) position of the corresponding DSAP-image. $\lfloor x \rfloor$ is the function to get the greatest integer less than or equal to $x$ and we set $S$ to 32 in our experiments.

	\textbf{Discussion.} For the DSAP-images of the $i$-th body part, the semantic identities on the same spatial positions are always the same. They are densely semantically aligned. 
	
	Such representations have three major advantages. 1) It overcomes spatial misalignment challenges resulting from diverse viewpoints and poses, and imperfect person detection. 2) It avoids the interference from diverse background clutters since only human body regions are warped to DSAP-images. 3) It is free from the appearance interference from occlusion since the semantics are not estimated over the occluding objects. 
	
	DSAP-images, however, have three limitations with respect to its roles in the person re-ID task. 1) The valid contents on the DSAP-images are very sparse (see Fig.~\ref{fig:2d-3d}). As a 2D projection from the 3D surface, nearly half of the body regions are invisible on the 2D image and thus cannot be detected by DensePose. Besides, there are usually estimation errors, including missing detections, especially on images with low resolution and blurring artifacts. 2) The dense semantics estimator is not optimal. Since there is no labeled dense semantics for the re-ID datasets, we leverage the DensePose model trained on the COCO-DesenPose dataset. However, there is a gap between these datasets in resolutions, image quality, and pose distributions. 3) Since the background is removed, some discriminative contents, such as a red backpack, are also removed.

	\subsection{Joint Learning of Our Network} 
	
	Due to the sparsity of valid contents and potential semantics estimation errors on the DSAP-images (as discussed in subsection \ref{subsec:SAPIC}), it is very challenging to design an effective network to exploit the semantically aligned information from the DSAP-images alone. In fact, a few of our early attempts along this line (with only the DSAP-images as input) have failed to deliver good results. To exploit the merits of the DSAP-images while addressing the above mentioned challenges, in our design, we propose treating them as regulators in an end-to-end network to drive the semantically aligned feature learning from the original full image. One important advantage of this design is that during the inference, the regulators are not needed, making it computationally efficient. This also removes the dependency of the inference on the performance of the dense semantics
	estimator, making the system practically more robust.
	
	Fig.~\ref{fig:flowchat-learning} shows the flowchat. The \emph{DSAG-Stream} plays the role of a regulator to assist the training of the \emph{MF-Stream}. We achieve this through the corresponding feature fusion between the \emph{DSAG-Stream} and the \emph{MF-Stream}, and the supervision on the fused features. For the \emph{DSAG-Stream}, the input DSAP-images are densely semantically aligned and thus the output features inherit the merits. We intend to leverage the \emph{DSAG-Stream} to drive the \emph{MF-Stream} to learn both global features and part-aware features. For each stream, a small head network with two branches are designed to focus on global and local information respectively.
	
	
	
	\subsubsection{DSAG-Stream}
	\vspace{-1mm}
	
	The \emph{DSAG-Stream} consists of the multi-branch sub-networks (MB-Ns) and a small Head network formed by a global branch and a part branch as shown in Fig.~\ref{fig:flowchat-learning}. We  show the detailed architectures in Table \ref{tab:DSAP-stream}.
	
	\textbf{Multi-branch sub-networks (MB-Ns)}. Both global information and local details are important and complementary for re-ID \cite{bai2017deep,wang2018learning,li2017person,wei2017glad,su2017pose,zhao2017spindle}. In order to learn local  detailed features of the separate region part rather than mixing all parts together, we adopt multi-branch sub-networks (MB-Ns) to learn local feature maps $\mathbf{D}_i \in \mathbb{R}^{h\times w \times c}$ of size $h\times w$ with $c$ channels, $i = 1,2,\cdots,N$, for $N$ merged body part regions, respectively. Note that the $N$ body part regions have no overlap. The $N$ feature maps are concatenated along channels and we have $\mathbf{D} = \mathbf{D}_i|_{i=1}^N = [\mathbf{D}_1, \mathbf{D}_2, \cdots, \mathbf{D}_N] \in \mathbb{R}^{h\times w \times c_A}$, where $c_A = N\!\times\!c$.
	
\begin{figure*}[th]
	\begin{center}
		\includegraphics[width=0.96\linewidth]{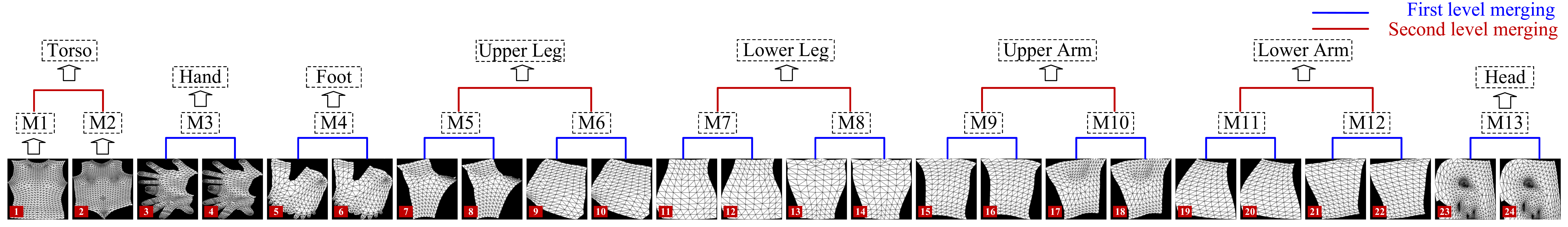}
	\end{center}
	\vspace{-1mm}
	\caption{Illustration of two level merging corresponding to the 24 body parts.}
	\label{fig:Merging}
    \end{figure*}

	For the MB-Ns, we have two levels of merging to progressively merge features from correlated body parts, in order to exploit the symmetry of human body to be better viewpoint robust and to reduce the number of branches. We obtain 8 separate feature maps from MB-Ns as $\mathbf{D}_i|_{i=1}^N$, where $N=8$. The semantics for a pair of left-right symmetric parts, are semantically aligned in the UV space and we element-wisely add the features in the first level merging. At the second level merging, similarly, we merge the two branches corresponding to the front-back symmetric parts and finally obtain 8 branches as illustrated in Fig.~\ref{fig:Merging}. 
	
	\textbf{Head network}. It consists of two separate branches which focus on global and local information respectively. 
	
	For the global branch, the output feature vector $\mathbf{d}_G\in \mathbb{R}^{2048}$ are obtained by
	\begin{equation}\label{eq:global}
		\mathbf{d}_G = \mathcal{P}(\mathcal{H}(\mathbf{D})), 
	\end{equation}
	where $\mathcal{H}(\mathbf{\cdot})$ denotes an underlying mapping consisting of a few stacked layers; $\mathcal{P}(\mathbf{\cdot})$ denotes the average spatial pooling operation. We take the network architecture of conv5\_g as shown in Table \ref{tab:DSAP-stream} for this mapping of $\mathcal{P}(\mathcal{H}(\mathbf{\cdot}))$. 
	
	For the part branch, the output feature vector $\mathbf{d}_L \!\in\! \mathbb{R}^{2048}$ is a concatenation of the feature vectors $\mathbf{d}_{L,i} \in \mathbb{R}^{256}$ of the 8 merged parts, {\it{i.e.}}, $\mathbf{d}_L = [\mathbf{d}_{L,1}, \mathbf{d}_{L,2}, \cdots, \mathbf{d}_{L,8}]$, with $\mathbf{d}_{L,i}$ obtained by
	\begin{equation}\label{eq:local_D}
		\mathbf{d}_{L,i} = \mathcal{P}(\mathcal{F}(\mathbf{\mathbf{D}_{i}})),  
	\end{equation}
	where $\mathcal{F}(\mathbf{X})$ denotes an underlying mapping consisting of a few stacked layers. We take the network architecture of conv5\_l as shown in Table~\ref{tab:DSAP-stream} for this mapping of $\mathcal{P}(\mathcal{F}(\mathbf{\cdot}))$. 
	\vspace{-3mm}
	\subsubsection{MF-Stream}
	\vspace{-1mm}
	We use the sub-network of ResNet-50 (conv1, conv2\_x, conv3\_x, and conv4\_x) \cite{he2016deep} to get the feature map $\mathbf{F}\in \mathbb{R}^{h\times w \times c_A} $. To facilitate the joint learning with the corresponding features from the \emph{DSAG-Stream}, with the feature map $\mathbf{F}$ as input (see Fig.~\ref{fig:flowchat-learning}), the global features $\mathbf{f}_{G}$ and the part-aware features $\mathbf{f}_{L}$ are learned by the two separate branches of a small Head network. This Head network architecture is similar to the Head network of the \emph{DSAG-Stream}. Note that the features $\mathbf{f}_{G}$ and $\mathbf{f}_{L}$ are used for re-ID in our final scheme. 
	
	
	\vspace{-3mm}
	\subsubsection{Two-Stream Fusion}
	\vspace{-1mm}
	We fuse the global features from the two streams by element-wise adding, {\it{i.e.}}, $\mathbf{z}_{G} = \mathbf{f}_{G} + \mathbf{d}_{G}$, which enables the joint optimization of the two streams. Similarly, the part-aware features and part features from the two streams are fused as $\mathbf{z}_{L} = \mathbf{f}_{L} + \mathbf{d}_{L}$.
	
	\vspace{-3mm}
	\subsubsection{Loss Designs}
	\vspace{-1mm}
		
	To train the network, we use the widely-used identification loss (ID Loss), {\it{i.e.}}, the cross entropy loss for identification classification, and the ranking loss of triplet loss with batch hard mining \cite{hermans2017defense} (Triplet Loss) as our loss functions. 
	
	Considering the noises in the DSAP-images due to semantics estimation errors, and the high complexity of the DensePose model, in our design, we treat the DSAP-images as regulators to drive the semantically aligned feature learning from the original full image, expecting the \emph{MF-Stream} alone to work in inference. We add supervision to the features $\mathbf{f}_{G}$, $\mathbf{f}_{L}$ from the \emph{MF-Stream} and to the fused features $\mathbf{z}_{G}$, $\mathbf{z}_{L}$, respectively as illustrated in Fig.~\ref{fig:flowchat-learning}. Specifically, for the \emph{MF-Stream}, we add the ID loss for the global feature vector $\mathbf{f}_{G}$, and each part-aware feature vector $\mathbf{f}_{L,i}, i = 1, 2, \cdots, 8$. For the fused features $\mathbf{z}_{G}$, $\mathbf{z}_{L}$, both the ID loss and triplet loss are added. The loss computed using the fused features makes the gradient back-propagated to the \emph{MF-Stream} be also influenced by the \emph{DSAG-Stream} features, since they contribute to the fused features and the resulting loss. In this way, the \emph{DSAG-Stream} plays the role of regularization by impacting the feature learning of the MF-Stream in the training.
	
	To calculate each identification loss, a classifier constructed by two fully connected (FC) layers followed by a SoftMax function is applied to the feature vector to output the classification probability.

	\begin{table}[t]
		\centering
		\caption{Detailed architecture of our \emph{DSAG-Stream}. We construct it using  similar convolutional layers and building blocks as in ResNet-18 \cite{he2016deep}. For conv1, \emph{5$\times$5,32} denotes the convolutional kernel size is 5$\times$5 and output channel number is 32. Following the representation style in \cite{he2016deep}, building blocks are shown in brackets, with the numbers of blocks stacked. Downsampling is performed by conv3\_1 and conv4\_1 with a stride of 2. \emph{\#Bran.} denotes the number of sub-branches.}
		\footnotesize
		\begin{tabular}{c|c|c|c|c}
			\hline
			\multicolumn{2}{c|}{Layer name} & \multicolumn{1}{c|}{Parameters} & \multicolumn{1}{c|}{Output size} & \#Bran. \bigstrut\\
			\hline
			\multicolumn{1}{c|}{\multirow{8}[12]{*}{MB-Ns}} &  conv1  &  $5\times5, 32$  &  $32\times32$  &  24 \bigstrut[t]\\
			\cline{2-5}
			\multicolumn{1}{c|}{} &  conv2  &  $3\times3, 64$  & $32\times32$ & 24 \bigstrut[t]\\
			\cline{2-5}
			\multicolumn{1}{c|}{} & \multicolumn{1}{c|}{\multirow{2}[2]{*}{conv3\_x}} & \multicolumn{1}{c|}{\multirow{2}[2]{*}{$\begin{bmatrix}3\times3,64\\3\times3,64\end{bmatrix}\times2$}} & \multicolumn{1}{c|}{\multirow{2}[2]{*}{$16\times16$}} & \multirow{2}[2]{*}{24} \bigstrut\\
			\multicolumn{1}{c|}{} & \multicolumn{1}{c|}{} & \multicolumn{1}{c|}{} & \multicolumn{1}{c|}{} & \multicolumn{1}{c}{} \bigstrut[b]\\
			\cline{2-5}
			\multicolumn{1}{c|}{} &  merging  &  element-wise add & $16\times16$ & $24\!\!\rightarrow\!\!13$ \bigstrut[t]\\
			\cline{2-5}
			\multicolumn{1}{c|}{} & \multicolumn{1}{c|}{\multirow{2}[2]{*}{conv4\_x}} & \multicolumn{1}{c|}{\multirow{2}[2]{*}{$\begin{bmatrix}3\times3,128\\3\times3,128\end{bmatrix}\times2$}} & \multicolumn{1}{c|}{\multirow{2}[2]{*}{$8\times8$}} & \multicolumn{1}{c}{\multirow{2}[2]{*}{13}} \bigstrut\\
			\multicolumn{1}{c|}{} & \multicolumn{1}{c|}{} & \multicolumn{1}{c|}{} & \multicolumn{1}{c|}{} & \multicolumn{1}{c}{} \bigstrut[b]\\
			\cline{2-5}
			\multicolumn{1}{c|}{} &  merging  &  element-wise add & $8\times8$ & $13\!\!\rightarrow\!8$ \bigstrut[t]\\
			\hline
			\multicolumn{1}{c|}{\multirow{6}[8]{*}{Head}} & \multicolumn{1}{c|}{\multirow{3}[4]{*}{conv5\_g}} & \multicolumn{1}{c|}{\multirow{2}[2]{*}{$\begin{bmatrix}3\times3,2048\\3\times3,2048\end{bmatrix}\times2$}} & \multicolumn{1}{c|}{\multirow{2}[2]{*}{$8\times8$}} & \multicolumn{1}{c}{\multirow{3}[4]{*}{1}} \bigstrut\\
			\multicolumn{1}{c|}{} & \multicolumn{1}{c|}{} & \multicolumn{1}{c|}{} & \multicolumn{1}{c|}{} & \multicolumn{1}{c}{} \bigstrut[b]\\
			\cline{3-4}
			\multicolumn{1}{c|}{} & \multicolumn{1}{c|}{} & Average Pooling & $1\times1$ & \bigstrut[t]\\
			\cline{2-5}
			\multicolumn{1}{c|}{} & \multicolumn{1}{c|}{\multirow{3}[4]{*}{conv5\_l}} & \multicolumn{1}{c|}{\multirow{2}[2]{*}{$\begin{bmatrix}3\times3,256\\3\times3,256\end{bmatrix}\times2$}} & \multicolumn{1}{c|}{\multirow{2}[2]{*}{$8\times8$}} & \multicolumn{1}{c}{\multirow{3}[4]{*}{8}} \bigstrut\\
			\multicolumn{1}{c|}{} & \multicolumn{1}{c|}{} & \multicolumn{1}{c|}{} & \multicolumn{1}{c|}{} & \multicolumn{1}{c}{} \bigstrut[b]\\
			\cline{3-4}
			\multicolumn{1}{c|}{} & \multicolumn{1}{c|}{} & Average Pooling & $1\times1$ &  \bigstrut[t]\\
			\hline
		\end{tabular}%
		\label{tab:DSAP-stream}%
	\end{table}%

	\section{Experiments}
	
	\subsection{Datasets and Evaluation Metrics}
	
	\textbf{Market1501} \cite{zheng2015scalable} has 32,668 DPM-detected pedestrain image boxes of 1,501 identities, with 12,936 training, 3,368 query and 19,732 gallery images. 751 identities are used for training while the remaining 750 for testing.

	\textbf{CUHK03} \cite{li2014deepreid} consists of 1,467 pedestrians. This dataset provides both manually labeled bounding boxes from 14,096 images and DPM-detected bounding boxes from 14,097 images. We adopt the new training/testing protocol following \cite{zhong2017re, zheng2018pedestrian, he2018recognizing}. In this protocol, 767 identities are used for training and the remaining for testing.

	\textbf{CUHK01} \cite{li2012human} comprises 3884 images of 971 identities, captured in two disjoint camera views. We adopt the common experimental setting following  \cite{ahmed2015improved,cheng2016person,zhao2017deeply}. 
	
	
	\textbf{DukeMTMC-reID} \cite{zheng2017unlabeled} is a subset of Duke Dataset \cite{ristani2016performance} for image-based re-ID. We use the standard training/testing split and evaluation setting following \cite{zheng2017unlabeled,lin2017improving}. It contains 16,522 training images of 702 identities, 2,228 query images of the other 702 identities and 17,661 gallery images. 
	
	\textbf{Evaluation Metrics.} Following the common practices,  we use the cumulative matching characteristics (CMC) at Rank-1 (at least), Rank-5, Rank-10, and mean average precision (mAP) to evaluate the accuracy.   
	
	
	\subsection{Implementation Details}
	
	\textbf{Network settings.} We take ResNet-50 \cite{he2016deep} to build our baseline networks as in some re-ID systems \cite{bai2017deep,sun2017beyond,zhang2017alignedreid,almazan2018re}. Similar to \cite{sun2017beyond}, the last spatial down-sample operation in the Conv5 layer is removed. 
	
	
	For the \emph{MF-Stream}, we use a part of the ResNet-50 architecture ({\it{i.e.}}, conv1, conv2\_x to conv4\_x) as the sub-network to obtain the feature map $\mathbf{F}$. The weights pretrained on ImageNet \cite{deng2009imagenet} are used for initialization. The Head network architecture is similar to the Head network of the \emph{DSAG-Stream} and is randomly initialized. The difference is that the architecture of the global branch in the \emph{MF-Stream} is the same as the network architecture of the conv5\_x block in ResNet-50 rather than that in ResNet-18. Each local branch of the \emph{MF-Stream} uses an architecture similar to the global branch but has only 1/8 of the number of channels on each layer. For the \emph{DSAG-Stream}, the network is randomly initialized and trained from scratch.

	
	
	\textbf{Data augmentation.} We use the commonly used data augmentation strategies of random cropping \cite{wang2018resource}, horizontal flipping and random erasing \cite{zhong2017random,wang2018resource,wang2018mancs} (with a probability of 0.5) in both the baseline schemes and our schemes.
	
	
	\textbf{Optimization.} For the triplet loss with batch hard mining \cite{hermans2017defense}, we sample $P=16$ identities and $K=4$ images \cite{wang2018learning} for each identity as a mini-batch and the margin parameter is set to 0.3. The ID loss for the \emph{MF-Stream} features, the triplet loss, and ID loss for the fused features are weighed by 0.5, 1.5 and 1.0 respectively. We adopt Adam optimizer with a weight decay of $5\times10^{-4}$ to train our models. We warm up the models for 20 epochs with a linear growth learning rate from $8\times10^{-6}$ to $8\times10^{-4}$. Then, the learning rate is decayed by a factor of 0.5 for every 40 epochs. We observe that the models converge after training of 320 epochs. All our models are implemented on PyTorch and trained in an end-to-end manner.

	\subsection{Comparison with  State-of-the-Art}

	\begin{table*}[t] 
		\centering
		\newcommand{\tabincell}[2]{\begin{tabular}{@{}#1@{}}#2\end{tabular}}
		\caption{Performance (\%) comparisons with the state of the art methods. Bold numbers denote the best performance, while numbers with underlines denote the second best. Superscript $*$ indicates that model is pre-trained on CUHK03 and fine-tuned on CUHK01.}
		\footnotesize 
		
		\resizebox{\textwidth}{!}{
			\begin{tabular}{crcccccccccc}
				\hline
				\multicolumn{2}{c}{\multirow{3}[6]{*}{Method}} & \multicolumn{2}{c}{\multirow{2}[4]{*}{Market1501 (SQ)}} & \multicolumn{4}{c}{CUHK03}& \multicolumn{2}{c}{\multirow{2}[4]{*}{CUHK01}} & \multicolumn{2}{c}{\multirow{2}[4]{*}{DukeMTMC-reID}} \bigstrut\\
				\cline{5-8}    \multicolumn{2}{c}{} & \multicolumn{2}{c}{} & \multicolumn{2}{c}{Labeled} & \multicolumn{2}{c}{Detected} & \multicolumn{2}{c}{} \bigstrut\\
				\cline{3-12} 
				\multicolumn{2}{c}{} & Rank-1 & mAP & Rank-1 & mAP & Rank-1 & mAP & Rank-1 & Rank-5 & Rank-1 & mAP \bigstrut\\
				\hline
				\multirow{2}[2]{*}{\tabincell{l}{Basic-CNN \\ (ResNet-50)}} & \multicolumn{1}{l}{IDE(ECCV18)~\cite{sun2017beyond}} & 85.3  & 68.5  & 43.8  & 38.9  & -  & - & -  & - & 73.2  & 52.8  \bigstrut[t]\\
				& \multicolumn{1}{l}{Gp-reid(Arxiv18)~\cite{almazan2018re}} & 92.2  & 81.2  & -  & -  & -  & - & -  & - & 85.2  & 72.8  \bigstrut[b]\\
				\hline
				\multirow{13}[2]{*}{\tabincell{l}{Pose/Part \\ -related}} & \multicolumn{1}{l}{Spindle(CVPR17)~\cite{zhao2017spindle}} & 76.9  & -  & -  & -  & -  & -  & 79.9  & 94.4  & -  & -  \bigstrut[t]\\
				& \multicolumn{1}{l}{PIE(Arxiv17)~\cite{zheng2017pose}} & 78.7  & 53.9  & -  & -  & -  & -  & -  & -  & -  & - \\
				& \multicolumn{1}{l}{MSCAN(CVPR17)~\cite{li2017learning}} & 80.8  & 57.5  & -  & -  & -  & -  & -  & -  & -  & - \\
				& \multicolumn{1}{l}{PDC(ICCV17)~\cite{su2017pose}} & 84.1  & 63.4  & -  & -  & -  & -  & -  & -  & -  & - \\
				& \multicolumn{1}{l}{Pose Transfer(CVPR18)~\cite{liu2018pose}} & 87.7  & 68.9  & 33.8  & 30.5  & 30.1  & 28.2 & -  & -  & 68.6  & 48.1 \\
				& \multicolumn{1}{l}{PN-GAN(ECCV18)~\cite{qian2018pose}} & 89.4  & 72.6  & -  & -  & -  & - & -  & -  & 73.6  & 53.2 \\
				& \multicolumn{1}{l}{PSE(CVPR18)~\cite{sarfraz2017pose}} & 87.7  & 69.0  & -  & -  & 30.2  & 27.3  & 67.7  & 86.6  & 79.8  & 62.0 \\
				& \multicolumn{1}{l}{MGCAM(CVPR18)~\cite{song2018mask}} & 83.8  & 74.3  & 50.1  & 50.2  & 46.7  & 46.9 & -  & -  & -  & - \\
				& \multicolumn{1}{l}{MaskReID(Arxiv18)~\cite{qi2018maskreid}} & 90.0  & 75.3  & -  & -  & -  & - & 84.3  & -  & 78.9  & 61.9 \\
				& \multicolumn{1}{l}{Part-Aligned(ECCV18)~\cite{suh2018part}} & 91.7  & 79.6  & -  & -  & -  & - & 80.7$^{*}$  & 94.4$^{*}$  & 84.4  & 69.3 \\    	
				& \multicolumn{1}{l}{AACN(CVPR18)~\cite{xu2018attention}} & 85.9  & 66.9 & -  & -  & -  & - & \underline{88.1}  & \underline{96.7}  & 76.8  & 59.3 \\    	
				& \multicolumn{1}{l}{SPReID(CVPR18)~\cite{kalayeh2018human}} & 92.5  & 81.3  & -  & -  & -  & - & -  & -  & 84.4  & 71.0 \bigstrut[b]\\   	
				\hline
				\multirow{4}[2]{*}{\tabincell{l}{Stripe \\ -based}} & \multicolumn{1}{l}{AlignedReID(Arxiv17)~\cite{zhang2017alignedreid}} & 91.8  & 79.3  & -  & -  & -  & -  & -  & -  & -  & - \bigstrut[t]\\
				& \multicolumn{1}{l}{Deep-Person(Arxiv17)~\cite{bai2017deep}} & 92.3  & 79.6  & -  & -  & -  & - & -  & -  & 80.9  & 64.8 \\
				& \multicolumn{1}{l}{PCB+RPP(ECCV18)~\cite{sun2017beyond}} & \underline{93.8}  & 81.6  & 63.7  & 57.5  & -  & - & -  & -  & 83.3  & 69.2 \\
				& \multicolumn{1}{l}{MGN(MM18)~\cite{wang2018learning}} & \textbf{95.7 } & \underline{86.9}  & \underline{68.0}  & \underline{67.4}  & \underline{66.8}  & \underline{66.0}  & -  & -  & \textbf{88.7 } & \textbf{78.4 } \bigstrut[b]\\
				
				\hline
				\multirow{4}[2]{*}{\tabincell{l}{Attention \\ -based}} & \multicolumn{1}{l}{DLPAP(ICCV17)~\cite{zhao2017deeply}} & 81.0  & 63.4  & -  & -  & -  & -  & 76.5$^{*}$  & 94.2$^{*}$  & -  & - \bigstrut[t]\\
				& \multicolumn{1}{l}{HA-CNN(CVPR18)~\cite{li2018harmonious}} & 91.2  & 75.7 & 44.4  & 41.0  & 41.7  & 38.6  & -  & -  & 80.5  & 63.8 \\ 
				& \multicolumn{1}{l}{DuATM(CVPR18)~\cite{si2018dual}} & 91.4  & 76.6  & -  & -  & -  & - & -  & -  & 81.8  & 64.6 \\
				& \multicolumn{1}{l}{Mancs(ECCV18)~\cite{wang2018mancs}} & 93.1  & 82.3  & 69.0  & 63.9  & 65.5  & 60.5 & -  & -  & 84.9  & 71.8 \bigstrut[b]\\
				\hline
				\tabincell{l}{Dense Semantics \\ -based (Ours)} & \multicolumn{1}{l}{DSA-reID} & \textbf{95.7 } & \textbf{87.6 } & \textbf{78.9 } & \textbf{75.2 } & \textbf{78.2 } & \textbf{73.1 } & \textbf{90.4}$^{*}$  & \textbf{97.8}$^{*}$  & \underline{86.2 }  & \underline{74.3 } \bigstrut\\
				\hline
			\end{tabular} }%
			\label{tab:SOTA}%
		\end{table*}%

		We compare our proposed Densely Semantically Aligned re-ID scheme (DSA-reID) with current state-of-the-art methods of four categories in Table \ref{tab:SOTA}. \emph{Basic-CNN methods} have similar network structures with a commonly used baseline in deep re-ID systems \cite{zheng2017pose,zhang2017alignedreid,bai2017deep,wang2018learning,wang2018mancs}, which learns a global descriptor. 
		\emph{Pose/Part-related methods} leverage the coarse pose/part semantic information to assist re-ID. \emph{Stripe-based methods} divide the full RGB image/feature map into several horizontal stripes to exploit local details. MGN~\cite{wang2018learning} combines the local features of multiple granularities and the global features. \emph{Attention-based methods} \cite{zhao2017deeply,li2018harmonious,si2018dual,wang2018mancs} jointly learn attention selection and feature representation. Note that we do not implement re-ranking \cite{zhong2017re} in all our models for clear comparisons.

		\textbf{Market-1501}.~DSA-reID achieves the best performance. Our method and the second best method MGN~\cite{wang2018learning} have similar performance and both outperform the other methods by at least +1.9\%/+4.6\% in Rank-1/mAP accuracy. We only show the single query (SQ) results to save space, and a similar trend is observed for the multiple query setting.  
		
		\textbf{CUHK03}. DSA-reID outperforms others by a large margin, at least \textbf{+10.9\%/+7.8\%} in Rank-1/mAP for the labeled setting, and \textbf{+11.4\%/+7.1\%} in Rank-1/mAP for the detected setting. The images are less blurred than those in other datasets. The semantics estimation is more accurate which greatly helps the training of our networks. 
		
		\textbf{CUHK01}. Our method outperforms the current best result by +2.3\%/+1.1\% in Rank-1/Rank-5 accuracy. Similar to the methods in~\cite{cheng2016person, zhao2017deeply, suh2018part}, this result is obtained with pre-training on CUHK03 and fine-tuning on CUHK01. For fair comparisons, we also test our model without a pre-training on CUHK03, it achieves 88.6\%/97.1\% in Rank-1/Rank-5 respectively, which are also the best.
		
		\textbf{DukeMTMC-reID}. DSA-reID achieves the second best results. The semantics estimation on this dataset is error prone. More than 20\% persons cannot be detected on the training images. DSA-reID outperforms all the other approaches except MGN~\cite{wang2018learning} which ensembles local features at multiple granularities. We believe training a better DensePose estimator can further improve the performance.

		\subsection{Ablation Study}
		
		We perform comprehensive ablation studies on the Market-1501 dataset (single query).
		
		\textbf{Ours vs. baselines}. In Table \ref{tab:ablation-baseline}, ``Baseline" and ``Baseline(RE)" denote our baseline schemes without and with random erasing (RE) \cite{almazan2018re,wang2018mancs}, respectively. Label smoothing regularization \cite{szegedy2016rethinking}, which acts as a mechanism to regularize the classifier layer by changing the ground-truth label distribution, has been demonstrated to be effective in recognition \cite{pereyra2017regularizing,xie2016disturblabel}. We add label smoothing (LS) to the classification sub-task in the re-ID and denote this baseline as ``Baseline(RE+LS)". It improves the Rank-1/mAP accuracy over ``Baseline(RE)" by +1.1\%/+2.6\%. Besides, we also take our \emph{MF-Stream} only scheme that is built based on ``Baseline(RE+LS)" but with a Head network of two branches as our baseline, referred to as ``Baseline (Two branches)".
		
		
		We denote the proposed densely semantically aligned (DSA) re-ID schemes under different settings/designs with the prefix of ``DSA". ``DSA(Two streams fused)" denotes our two stream scheme which takes $\mathbf{z}_G$ and $\mathbf{z}_L$ as the matching features for inference. In inference, the \emph{DSAG-Stream} can be discarded and we refer to it as ``DSA-reID(Only \emph{MF-Stream})", which takes $\mathbf{f}_G$ and $\mathbf{f}_L$ as the matching features and is our final scheme, also named as ``DSA-reID". 
		
		We have the following observations/conclusions. 1) Our final scheme achieves significant performance improvement, outperforming ``Baseline (RE+LS)" by \textbf{+2.3\%/+6.4\%} and ``Baseline(Two branches)" by \textbf{+1.7\%/+4.2\%} in Rank-1/mAP accuracy respectively. 2) ``DSA-reID(Only \emph{MF-Stream})" has very similar performance as ``DSA(Two streams fused)" but much lower computational complexity.

		\begin{table}[t]
			\centering
			\caption{Performance (\%) comparisons of baselines and our schemes on the Market-1501 dataset.}
			\footnotesize
			\tabcolsep=5pt
			\begin{tabular}{ccccc}
				\hline
				Model  & mAP & Rank-1 & Rank-5 & Rank-10 \bigstrut\\
				\hline
				Baseline  & 76.4 & 91.2 & 96.5 & 97.9 \bigstrut[t]\\
				Baseline (RE) & 78.6 & 92.3 & 97.6 & 98.3 \bigstrut[t]\\
				Baseline (RE+LS)  & 81.2 & 93.4 & 97.8 & 98.5 \bigstrut[t]\\
				Baseline (Two branches)  & 83.4 & 94.0 & 98.0 & 98.7 \bigstrut[t]\\
				\hline
				DSA-Global(Single) & 84.7 & 94.8 & 98.2 & 98.9 \bigstrut[t]\\
				DSA-Local(Single) & 83.2 & 94.0 & 97.9 & 98.6 \bigstrut[t]\\
				\hline
				DSA-Global(Joint) & 87.4 & 95.6 & 98.6 & 99.1 \bigstrut[t]\\
				DSA-Local(Joint) & 86.5 & 95.2 & 98.4 & 99.0 \bigstrut[t]\\
				DSA(Two streams fused) & 87.5 & \bf{95.8} & \bf{98.4} & \bf{99.1} \bigstrut[t]\\
				\hline
				DSA-reID(Only MF-Stream) & \bf{87.6} & 95.7 & \bf{98.4} & \bf{99.1} \bigstrut[t]\\		
				\hline
			\end{tabular}%
			\label{tab:ablation-baseline}%
			\vspace{-3mm}
		\end{table}%
		
		\textbf{Global and part-aware/part features}. For each stream, we have two branches which focus on global features and part features resepctively. We show the analysis in Table \ref{tab:ablation-baseline}. 1) ``DSA-Global(Single)"/``DSA-Local(Single)" denotes the design with only the global/local branch in our two stream framework for both training and inferencing. ``DSA-Global(Single)" outperforms ``Baseline(RE+LS)" by +1.4\%/+3.5\% in Rank-1/mAP accuracy. ``DSA-Local(Single)" outperforms ``Baseline(RE+LS)" by +0.6\%/+2.0\% in Rank-1/mAP accuracy. These demonstrate that our semantic alignment design is very efficient. 2) Since global and part-aware/part features are complementary, our scheme with both the global and part-aware/part branches, ``DSA(Two streams fused)", achieves additional +1.0\%/+2.8\%, and +1.8\%/+4.3\% gain in comparison with ``DSA-Global(Single)" and ``DSA-Local(Single)" in Rank-1/mAP accuracy respectively. 3) ``DSA-Global(Joint)" or ``DSA-Local(Joint)" denotes that the inference is based on the features of the global branch or part-aware branch of our scheme ``DSA(Two stream fused)", {\it{i.e.}}, $\mathbf{z}_G$ or $\mathbf{z}_L$. Thanks to the joint training, ``DSA-Global/Local(Joint)" significantly outperforms ``DSA-Global/Local(Single)".

		\textbf{Dense vs. coarse semantic alignment}. Thanks to the densely semantically aligned representation and our architecture design, our scheme achieves excellent performance. We take the DSAP-images as input to the \emph{DSAG-Stream}. One may wonder about the performance if the cropped body parts without internal fine grained alignment are taken as input to our framework. We conduct an experiment by replacing the 24 DSAP-images by 24 cropped part images (without alignment within a part region) and refer to this scheme as coarsely semantically aligned re-ID, CSA. Table \ref{tab:coarse-fine} shows the performance comparisons. 1) Our densely semantically aligned scheme significantly outperforms the coarsely semantically aligned scheme by +1.6\%/+3.5\% in Rank-1/mAP accuracy. 2) Our coarsely semantically aligned scheme still outperforms the baselines by a large margin, demonstrating the effectiveness of our architecture design. 
		
		\begin{table}[t]
			\centering
			\caption{Performance (\%) comparisons between dense and coarse semantic alignment in our framework on the Market-1501 dataset.}
			\footnotesize
			\begin{tabular}{ccccc}
				\hline
				Model & mAP & Rank-1 & Rank-5 & Rank-10 \bigstrut\\
				\hline
				Baseline (RE+LS)  & 81.2 & 93.4 & 97.8 & 98.5 \bigstrut[t]\\
				\hline
				CSA(Only \emph{MF-Stream}) & 84.1 & 94.1 & 98.1 & 98.8 \bigstrut[t]\\
				DSA(Only \emph{MF-Stream})  & \bf{87.6} & \bf{95.7} & \bf{98.4} & \bf{99.1} \bigstrut[t]\\
				\hline
			\end{tabular}%
			\label{tab:coarse-fine}%
			\vspace{-2mm}
		\end{table}%
		\textbf{Two stream fusion designs}. We investigate how to make the \emph{MF-Stream} and the \emph{DSAG-Stream} interact efficiently for joint training and show the comparisons in Table \ref{tab:fusion}.  ``Concatenation+fc" denotes that for either branch, the features from the \emph{MF-Stream} and the \emph{DSAG-Stream} are concatenated followed by a fully connected layer. ``Elem-add" denotes that the features from the \emph{MF-Stream} and the \emph{DSAG-Stream}  are element-wisely added. ``Concatination+fc" has poor performance. In contrast, our fusion with element-wise add achieves excellent performance. 
		
		
		
		\begin{table}[htbp]
			\centering
			\caption{Performance (\%) comparisons on the designs of the two-stream fusion, on the Market-1501 dataset.}
			\footnotesize
			\tabcolsep=8pt
			\begin{tabular}{ccccc}
				\hline
				Fusion method & mAP & Rank-1 & Rank-5 & Rank-10 \bigstrut[t]\\
				\hline
				Concatenation+fc & 81.6 & 93.0 & 97.6 & 98.6 \bigstrut[t]\\
				Elem-add & \bf{87.6} & \bf{95.7} & \bf{98.4} & \bf{99.1} \bigstrut[t]\\
				\hline
			\end{tabular}%
			\label{tab:fusion}%
			\vspace{-5mm}
		\end{table}%

		
		
		\section{Conclusion}
		In this paper, we propose a densely semantically aligned person re-ID framework, intending to address the ubiquitous misalignment problems. Thanks to the estimated dense semantics, it becomes possible to construct the densely semantically aligned part images (DSAP-images) from the 2D image. We design a two stream network consisting of the \emph{MF-Stream} and the \emph{DSAG-Stream}. Considering that the DSAP-images have the inherent densely semantically aligned merits, but are noisy due to semantics estimation error, we treat the \emph{DSAG-Stream} as a regulator to assist the feature learning of the \emph{MF-Stream}, through our fusion and supervision designs. In the inference, only the \emph{MF-Stream} is needed, making the system more computationally efficient and robust. Our scheme achieves the best performance on Market-1501, CUHK03, and CUHK01. On CUHK03, our scheme significantly outperforms the previous methods, by at least \textbf{+10.9\%/+7.8\%} in Rank-1/mAP accuracy.

\section*{Acknowledgements}
This work was partly supported by the National Key Research and Development Program of China under Grant No. 2016YFC0801001, the National Program on Key Basic Research Projects (973 Program) under Grant 2015CB351803, NSFC under Grant 61571413, 61390514.

		{\small
			\bibliographystyle{ieee_fullname}
			\bibliography{main}

\begin{thebibliography}{10}\itemsep=-1pt

\bibitem{ahmed2015improved}
Ejaz Ahmed, Michael Jones, and Tim~K Marks.
\newblock An improved deep learning architecture for person re-identification.
\newblock In {\em CVPR}, 2015.

\bibitem{almazan2018re}
Jon Almazan, Bojana Gajic, Naila Murray, and Diane Larlus.
\newblock Re-id done right: towards good practices for person
  re-identification.
\newblock {\em arXiv preprint arXiv:1801.05339}, 2018.

\bibitem{bai2017deep}
Xiang Bai, Mingkun Yang, Tengteng Huang, Zhiyong Dou, Rui Yu, and Yongchao Xu.
\newblock Deep-person: Learning discriminative deep features for person
  re-identification.
\newblock {\em arXiv preprint arXiv:1711.10658}, 2017.

\bibitem{chen2017multi}
Weihua Chen, Xiaotang Chen, Jianguo Zhang, and Kaiqi Huang.
\newblock A multi-task deep network for person re-identification.
\newblock In {\em AAAI}, 2017.

\bibitem{cheng2016person}
De Cheng, Yihong Gong, Sanping Zhou, Jinjun Wang, and Nanning Zheng.
\newblock Person re-identification by multi-channel parts-based cnn with
  improved triplet loss function.
\newblock In {\em CVPR}, 2016.

\bibitem{cheng2011custom}
Dong~Seon Cheng, Marco Cristani, Michele Stoppa, Loris Bazzani, and Vittorio
  Murino.
\newblock Custom pictorial structures for re-identification.
\newblock In {\em BMVC}, 2011.

\bibitem{deng2009imagenet}
Jia Deng, Wei Dong, Richard Socher, Li-Jia Li, Kai Li, and Li Fei-Fei.
\newblock Imagenet: A large-scale hierarchical image database.
\newblock In {\em CVPR}, 2009.

\bibitem{geng2016deep}
Mengyue Geng, Yaowei Wang, Tao Xiang, and Yonghong Tian.
\newblock Deep transfer learning for person re-identification.
\newblock {\em arXiv preprint arXiv:1611.05244}, 2016.

\bibitem{guler2018densepose}
R{\i}za~Alp G{\"u}ler, Natalia Neverova, and Iasonas Kokkinos.
\newblock Dense{P}ose: Dense human pose estimation in the wild.
\newblock {\em CVPR}, 2018.

\bibitem{guler2017densereg}
Riza~Alp G{\"u}ler, George Trigeorgis, Epameinondas Antonakos, Patrick Snape,
  Stefanos Zafeiriou, and Iasonas Kokkinos.
\newblock Densereg: Fully convolutional dense shape regression in-the-wild.
\newblock In {\em CVPR}, 2017.

\bibitem{he2016deep}
Kaiming He, Xiangyu Zhang, Shaoqing Ren, and Jian Sun.
\newblock Deep residual learning for image recognition.
\newblock In {\em CVPR}, 2016.

\bibitem{he2018recognizing}
Lingxiao He, Zhenan Sun, Yuhao Zhu, and Yunbo Wang.
\newblock Recognizing partial biometric patterns.
\newblock {\em arXiv preprint arXiv:1810.07399}, 2018.

\bibitem{hermans2017defense}
Alexander Hermans, Lucas Beyer, and Bastian Leibe.
\newblock In defense of the triplet loss for person re-identification.
\newblock {\em arXiv preprint arXiv:1703.07737}, 2017.

\bibitem{kalayeh2018human}
Mahdi~M Kalayeh, Emrah Basaran, Muhittin G{\"o}kmen, Mustafa~E Kamasak, and
  Mubarak Shah.
\newblock Human semantic parsing for person re-identification.
\newblock In {\em CVPR}, 2018.

\bibitem{li2017learning}
Dangwei Li, Xiaotang Chen, Zhang Zhang, and Kaiqi Huang.
\newblock Learning deep context-aware features over body and latent parts for
  person re-identification.
\newblock In {\em CVPR}, 2017.

\bibitem{li2012human}
Wei Li, Rui Zhao, and Xiaogang Wang.
\newblock Human reidentification with transferred metric learning.
\newblock In {\em ACCV}, 2012.

\bibitem{li2014deepreid}
Wei Li, Rui Zhao, Tong Xiao, and Xiaogang Wang.
\newblock Deepreid: Deep filter pairing neural network for person
  re-identification.
\newblock In {\em CVPR}, 2014.

\bibitem{li2017person}
Wei Li, Xiatian Zhu, and Shaogang Gong.
\newblock Person re-identification by deep joint learning of multi-loss
  classification.
\newblock {\em IJCAI}, 2017.

\bibitem{li2018harmonious}
Wei Li, Xiatian Zhu, and Shaogang Gong.
\newblock Harmonious attention network for person re-identification.
\newblock In {\em CVPR}, 2018.

\bibitem{lin2017improving}
Yutian Lin, Liang Zheng, Zhedong Zheng, Yu Wu, and Yi Yang.
\newblock Improving person re-identification by attribute and identity
  learning.
\newblock {\em arXiv preprint arXiv:1703.07220}, 2017.

\bibitem{liu2017end}
Hao Liu, Jiashi Feng, Meibin Qi, Jianguo Jiang, and Shuicheng Yan.
\newblock End-to-end comparative attention networks for person
  re-identification.
\newblock {\em TIP}, pages 3492--3506.

\bibitem{liu2018pose}
Jinxian Liu, Bingbing Ni, Yichao Yan, Peng Zhou, Shuo Cheng, and Jianguo Hu.
\newblock Pose transferrable person re-identification.
\newblock In {\em CVPR}, 2018.

\bibitem{oreifej2010human}
Omar Oreifej, Ramin Mehran, and Mubarak Shah.
\newblock Human identity recognition in aerial images.
\newblock In {\em CVPR}, 2010.

\bibitem{pereyra2017regularizing}
Gabriel Pereyra, George Tucker, Jan Chorowski, {\L}ukasz Kaiser, and Geoffrey
  Hinton.
\newblock Regularizing neural networks by penalizing confident output
  distributions.
\newblock {\em arXiv preprint arXiv:1701.06548}, 2017.

\bibitem{qi2018maskreid}
Lei Qi, Jing Huo, Lei Wang, Yinghuan Shi, and Yang Gao.
\newblock Maskreid: A mask based deep ranking neural network for person
  re-identification.
\newblock {\em arXiv preprint arXiv:1804.03864}, 2018.

\bibitem{qian2018pose}
Xuelin Qian, Yanwei Fu, Wenxuan Wang, Tao Xiang, Yang Wu, Yu-Gang Jiang, and
  Xiangyang Xue.
\newblock Pose-normalized image generation for person re-identification.
\newblock In {\em ECCV}, 2018.

\bibitem{ristani2016performance}
Ergys Ristani, Francesco Solera, Roger Zou, Rita Cucchiara, and Carlo Tomasi.
\newblock Performance measures and a data set for multi-target, multi-camera
  tracking.
\newblock In {\em ECCV}, 2016.

\bibitem{sarfraz2017pose}
M~Saquib Sarfraz, Arne Schumann, Andreas Eberle, and Rainer Stiefelhagen.
\newblock A pose-sensitive embedding for person re-identification with expanded
  cross neighborhood re-ranking.
\newblock In {\em CVPR}, 2018.

\bibitem{schroff2015facenet}
Florian Schroff, Dmitry Kalenichenko, and James Philbin.
\newblock Facenet: A unified embedding for face recognition and clustering.
\newblock In {\em CVPR}, 2015.

\bibitem{shen2015person}
Yang Shen, Weiyao Lin, Junchi Yan, Mingliang Xu, Jianxin Wu, and Jingdong Wang.
\newblock Person re-identification with correspondence structure learning.
\newblock In {\em ICCV}, 2015.

\bibitem{si2018dual}
Jianlou Si, Honggang Zhang, Chun-Guang Li, Jason Kuen, Xiangfei Kong, Alex~C
  Kot, and Gang Wang.
\newblock Dual attention matching network for context-aware feature sequence
  based person re-identification.
\newblock In {\em CVPR}, 2018.

\bibitem{song2018mask}
Chunfeng Song, Yan Huang, Wanli Ouyang, and Liang Wang.
\newblock Mask-guided contrastive attention model for person re-identification.
\newblock In {\em CVPR}, 2018.

\bibitem{su2017pose}
Chi Su, Jianing Li, Shiliang Zhang, Junliang Xing, Wen Gao, and Qi Tian.
\newblock Pose-driven deep convolutional model for person re-identification.
\newblock In {\em ICCV}, 2017.

\bibitem{suh2018part}
Yumin Suh, Jingdong Wang, Siyu Tang, Tao Mei, and Kyoung~Mu Lee.
\newblock Part-aligned bilinear representations for person re-identification.
\newblock In {\em ECCV}, 2018.

\bibitem{sun2017beyond}
Yifan Sun, Liang Zheng, Yi Yang, Qi Tian, and Shengjin Wang.
\newblock Beyond part models: Person retrieval with refined part pooling.
\newblock 2018.

\bibitem{szegedy2016rethinking}
Christian Szegedy, Vincent Vanhoucke, Sergey Ioffe, Jon Shlens, and Zbigniew
  Wojna.
\newblock Rethinking the inception architecture for computer vision.
\newblock In {\em CVPR}, 2016.

\bibitem{varior2016siamese}
Rahul~Rama Varior, Bing Shuai, Jiwen Lu, Dong Xu, and Gang Wang.
\newblock A siamese long short-term memory architecture for human
  re-identification.
\newblock In {\em ECCV}, 2016.

\bibitem{wang2018mancs}
Cheng Wang, Qian Zhang, Chang Huang, Wenyu Liu, and Xinggang Wang.
\newblock Mancs: A multi-task attentional network with curriculum sampling for
  person re-identification.
\newblock In {\em ECCV}, 2018.

\bibitem{wang2018learning}
Guanshuo Wang, Yufeng Yuan, Xiong Chen, Jiwei Li, and Xi Zhou.
\newblock Learning discriminative features with multiple granularities for
  person re-identification.
\newblock {\em ACM Multimedia}, 2018.

\bibitem{wang2013intelligent}
Xiaogang Wang.
\newblock Intelligent multi-camera video surveillance: A review.
\newblock {\em Pattern recognition letters}, 34(1):3--19, 2013.

\bibitem{wang2018resource}
Yan Wang, Lequn Wang, Yurong You, Xu Zou, Vincent Chen, Serena Li, Gao Huang,
  Bharath Hariharan, and Kilian~Q Weinberger.
\newblock Resource aware person re-identification across multiple resolutions.
\newblock In {\em CVPR}, 2018.

\bibitem{wei2017glad}
Longhui Wei, Shiliang Zhang, Hantao Yao, Wen Gao, and Qi Tian.
\newblock Glad: global-local-alignment descriptor for pedestrian retrieval.
\newblock In {\em ACM Multimedia}, pages 420--428, 2017.

\bibitem{wu2016personnet}
Lin Wu, Chunhua Shen, and Anton van~den Hengel.
\newblock Personnet: Person re-identification with deep convolutional neural
  networks.
\newblock {\em arXiv preprint arXiv:1601.07255}, 2016.

\bibitem{xie2016disturblabel}
Lingxi Xie, Jingdong Wang, Zhen Wei, Meng Wang, and Qi Tian.
\newblock Disturblabel: Regularizing cnn on the loss layer.
\newblock In {\em CVPR}, 2016.

\bibitem{xu2018attention}
Jing Xu, Rui Zhao, Feng Zhu, Huaming Wang, and Wanli Ouyang.
\newblock Attention-aware compositional network for person re-identification.
\newblock {\em arXiv preprint arXiv:1805.03344}, 2018.

\bibitem{xu2013human}
Yuanlu Xu, Liang Lin, Wei-Shi Zheng, and Xiaobai Liu.
\newblock Human re-identification by matching compositional template with
  cluster sampling.
\newblock In {\em ICCV}, 2013.

\bibitem{yao2017deep}
Hantao Yao, Shiliang Zhang, Yongdong Zhang, Jintao Li, and Qi Tian.
\newblock Deep representation learning with part loss for person
  re-identification.
\newblock {\em arXiv preprint arXiv:1707.00798}, 2017.

\bibitem{zhang2017alignedreid}
Xuan Zhang, Hao Luo, Xing Fan, Weilai Xiang, Yixiao Sun, Qiqi Xiao, Wei Jiang,
  Chi Zhang, and Jian Sun.
\newblock Alignedreid: Surpassing human-level performance in person
  re-identification.
\newblock {\em arXiv preprint arXiv:1711.08184}, 2017.

\bibitem{zhao2017spindle}
Haiyu Zhao, Maoqing Tian, Shuyang Sun, Jing Shao, Junjie Yan, Shuai Yi,
  Xiaogang Wang, and Xiaoou Tang.
\newblock Spindle net: Person re-identification with human body region guided
  feature decomposition and fusion.
\newblock In {\em CVPR}, 2017.

\bibitem{zhao2017deeply}
Liming Zhao, Xi Li, Yueting Zhuang, and Jingdong Wang.
\newblock Deeply-learned part-aligned representations for person
  re-identification.
\newblock In {\em ICCV}, pages 3239--3248, 2017.

\bibitem{zhao2013unsupervised}
Rui Zhao, Wanli Ouyang, and Xiaogang Wang.
\newblock Unsupervised salience learning for person re-identification.
\newblock In {\em CVPR}, 2013.

\bibitem{zhao2014learning}
Rui Zhao, Wanli Ouyang, and Xiaogang Wang.
\newblock Learning mid-level filters for person re-identification.
\newblock In {\em CVPR}, 2014.

\bibitem{zheng2017pose}
Liang Zheng, Yujia Huang, Huchuan Lu, and Yi Yang.
\newblock Pose invariant embedding for deep person re-identification.
\newblock {\em arXiv preprint arXiv:1701.07732}, 2017.

\bibitem{zheng2015scalable}
Liang Zheng, Liyue Shen, Lu Tian, Shengjin Wang, Jingdong Wang, and Qi Tian.
\newblock Scalable person re-identification: A benchmark.
\newblock In {\em ICCV}, 2015.

\bibitem{zheng2016person}
Liang Zheng, Yi Yang, and Alexander~G Hauptmann.
\newblock Person re-identification: Past, present and future.
\newblock {\em arXiv preprint arXiv:1610.02984}, 2016.

\bibitem{zheng2017unlabeled}
Zhedong Zheng, Liang Zheng, and Yi Yang.
\newblock Unlabeled samples generated by gan improve the person
  re-identification baseline in vitro.
\newblock {\em arXiv preprint arXiv:1701.07717}, 2017.

\bibitem{zheng2018pedestrian}
Zhedong Zheng, Liang Zheng, and Yi Yang.
\newblock Pedestrian alignment network for large-scale person
  re-identification.
\newblock {\em TCSVT}, 2018.

\bibitem{zhong2017re}
Zhun Zhong, Liang Zheng, Donglin Cao, and Shaozi Li.
\newblock Re-ranking person re-identification with k-reciprocal encoding.
\newblock In {\em CVPR}, 2017.

\bibitem{zhong2017random}
Zhun Zhong, Liang Zheng, Guoliang Kang, Shaozi Li, and Yi Yang.
\newblock Random erasing data augmentation.
\newblock {\em arXiv preprint arXiv:1708.04896}, 2017.

\end{thebibliography}
		}
		
	\end{document}